%% file: main.tex
\documentclass{article} %
\usepackage{colm2024_conference}

\usepackage{booktabs}
\usepackage{graphicx}
\usepackage{enumitem}
\usepackage{needspace}
\usepackage{wrapfig}
\usepackage{algorithm}
\usepackage{algpseudocode}
\usepackage{wrapfig}
\usepackage{float}
\usepackage{microtype}
\usepackage{amsmath}
\usepackage{amssymb}
\usepackage{mathtools}
\usepackage{colortbl}
\usepackage[utf8]{inputenc}
\usepackage{caption}
\usepackage{subcaption}
\usepackage{xcolor}
\usepackage{setspace}
\usepackage{url}
\usepackage{multirow}
\usepackage{colortbl}
\usepackage{tabularx}
\usepackage{arydshln}
\usepackage{blindtext}
\usepackage{pgfplots}
\pgfplotsset{compat=1.18}
\usepackage{tikz}
\usetikzlibrary{er, positioning, bayesnet}
\usepackage{makecell}
\usepackage{siunitx}
\usepackage{nicefrac}
\usepackage{tocloft}
\usepackage{listings}
\usepackage[raster, skins]{tcolorbox} %
\usepackage{xltabular}
\usepackage{adjustbox}
\usepackage{xurl}
\usepackage{xspace}
\usepackage{longtable}
\usepackage{pifont}
\input{math_commands.tex}

\definecolor{lightgray}{rgb}{0.9,0.9,0.9}

\renewcommand{\arraystretch}{1.25}

\makeatletter
\DeclareRobustCommand\onedot{\futurelet\@let@token\@onedot}
\def\@onedot{\ifx\@let@token.\else.\null\fi\xspace}

\makeatother

\def\hunyuanhome{\raisebox{-1.5pt}{\includegraphics[height=1.05em]{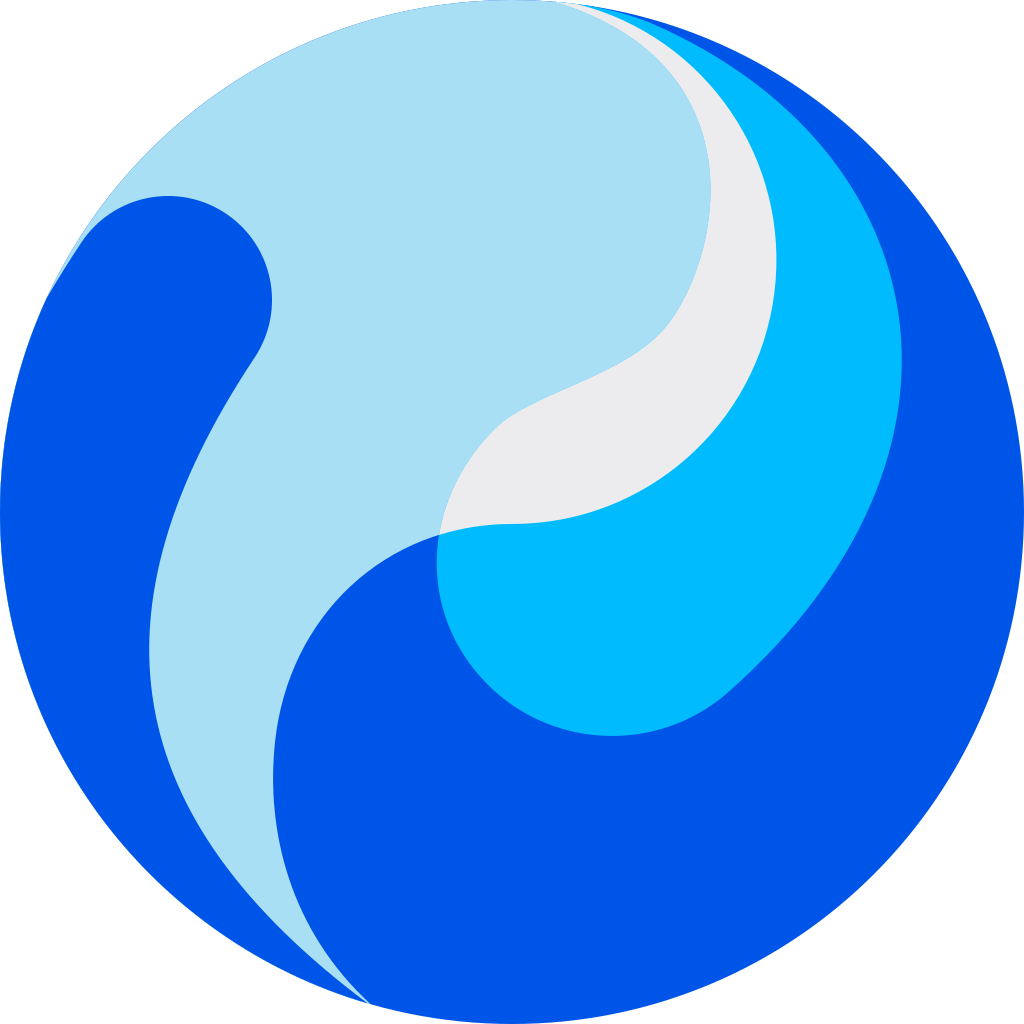}}}
\def\github{\raisebox{-1.5pt}{\includegraphics[height=1.05em]{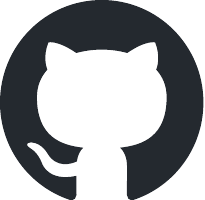}}}

\urldef{\homelinknew}\url{https://hy-soar.github.io}
\urldef{\ghlinknew}\url{https://github.com/Tencent-Hunyuan/HY-SOAR}

\title{
SOAR: Self-Correction for Optimal Alignment \\and Refinement in Diffusion Models
}

\author{
You Qin$^{*}$
\quad
Linqing Wang$^{*\dagger}$
\quad
Hao Fei
\quad
Roger Zimmermann
\quad
Liefeng Bo
\quad
Qinglin Lu
\quad
Chunyu Wang$^{\ddagger}$ 
\\
{\ttfamily\scriptsize * Equal contribution.
 $\dagger$ Project lead.
 $\ddagger$ Corresponding authors.}
}

\newcommand{\basemodel}{SD3.5-Medium\xspace}
\newcommand{\methodname}{SOAR\xspace}
\newcommand{\methodnamebf}{\textbf{SOAR}\xspace}

\newcommand{\cmark}{{\color{green!70!black}\ding{51}}}
\newcommand{\xmark}{{\color{red!80!black}\ding{55}}}
\makeatletter
\def\@maketitle{%
  \vbox{\hsize\textwidth
    {\centering {\Large\bf \@title\par}} 
    \vspace{-0.3cm}
    \begin{quote}\rule{\z@}{6pt} {\centering\par{\@author}\par}\end{quote}
    \vskip 0.18in minus 0.05in
  }%
  \begin{center}
    \vspace{-1cm} 
    \begin{tabular}{rl}
      \hunyuanhome & \homelinknew \\
      \github & \ghlinknew \\
    \end{tabular}
  \end{center}
  \thispagestyle{firstpage} 
}
\makeatother

\begin{document}
\maketitle

\input{sec/0_abstract}
\input{sec/1_introduction}
\input{sec/2_methodology}
\input{sec/3_experiment}
\input{sec/4_relatedwork}
\input{sec/5_conclusion}

\clearpage
\input{sec/x_contributor}
\clearpage

\appendix
\input{sec/a_appendix_diffusion_derivation}

\bibliography{colm2024_conference}
\bibliographystyle{colm2024_conference}
\end{document}

%% file: math_commands.tex
\usepackage{amsmath,amsfonts,bm}

\def\eqref#1{equation~\ref{#1}}

\def\1{\bm{1}}

\DeclareMathAlphabet{\mathsfit}{\encodingdefault}{\sfdefault}{m}{sl}
\SetMathAlphabet{\mathsfit}{bold}{\encodingdefault}{\sfdefault}{bx}{n}



%% file: sec/0_abstract.tex
\begin{abstract}
The post-training pipeline for diffusion models currently has two stages: supervised fine-tuning (SFT) on curated data and reinforcement learning (RL) with reward models. A fundamental gap separates them. SFT optimizes the denoiser only on ground-truth states sampled from the forward noising process; once inference deviates from these ideal states, subsequent denoising relies on out-of-distribution generalization rather than learned correction, exhibiting the same \emph{exposure bias} that afflicts autoregressive models, but accumulated along the denoising trajectory instead of the token sequence. RL can in principle address this mismatch, yet its terminal reward signal is sparse, suffers from credit-assignment difficulty, and risks reward hacking. We propose \methodnamebf (Self-Correction for Optimal Alignment and Refinement), a bias-correction post-training method that fills this gap. Starting from a real sample, \methodname performs a single stop-gradient rollout with the current model, re-noises the resulting off-trajectory state, and supervises the model to steer back toward the original clean target. The method is on-policy, reward-free, and provides dense per-timestep supervision with no credit-assignment problem. On SD3.5-Medium, \methodname improves GenEval from 0.70 to 0.78 and OCR from 0.64 to 0.67 over SFT, while simultaneously raising all model-based preference scores. In controlled reward-specific experiments, \methodname surpasses Flow-GRPO in final metric value on both aesthetic and text-image alignment tasks, despite having no access to a reward model. Since \methodname's base loss subsumes the standard SFT objective, it can directly replace SFT as a stronger first post-training stage after pretraining, while remaining fully compatible with subsequent RL alignment.
\end{abstract}

%% file: sec/1_introduction.tex
\section{Introduction}

Diffusion and flow-matching models have become the dominant paradigm for high-quality visual generation, powering text-to-image~\citep{LDM_rombach_2022,sd35_esser_2024,Flux_blacklabs_2025}, text-to-video~\citep{hunyuanvideo_kong_2024}, and 3D synthesis systems. Despite rapid progress in scaling and architecture, the \emph{post-training} recipe for these generators remains far less mature than its counterpart in large language models. In current practice a model is first pretrained at scale, then fine-tuned on curated data (SFT), and optionally aligned with reward-based optimization (RL). Yet even strong models continue to suffer from recurring failures: wrong object counts, malformed hands, broken text rendering, and unstable identity preservation.

We argue that these failures share a common root: \emph{exposure bias} in the denoising trajectory. In autoregressive language models, exposure bias refers to the mismatch between ground-truth-conditioned training and autoregressive inference~\citep{bengio_scheduled_2015}. Diffusion models suffer from the same phenomenon along a different axis. Standard SFT constructs noisy training states via the forward process of real data and optimizes the denoiser on these ideal, ground-truth states. During inference, however, the model conditions on its own earlier predictions. Once an early denoising step makes a small error, subsequent states enter regions that were not directly optimized; recovery then depends on out-of-distribution generalization rather than a learned correction mechanism. The error compounds over dozens of sampling steps and can become irreversible well before the final image is rendered.

A natural response is to apply preference-based or reward-based optimization. Offline methods such as Diffusion-DPO~\citep{wallace_diffusiondpo_2024} adapt direct preference optimization to diffusion models by learning from human-annotated image pairs, while online methods such as Flow-GRPO~\citep{flowgrpo_liu_2025} convert the deterministic ODE sampler into a stochastic SDE and apply policy-gradient exploration, achieving impressive improvements on compositional benchmarks. Similarly, DiffusionNFT~\citep{zheng_diffusionnft_2026} contrasts reward-split positive and negative generations on the forward process to define an implicit policy improvement direction. However, these approaches share fundamental challenges: the preference or reward signal is sparse and terminal, only available after a complete rollout, making credit assignment across dozens of denoising steps difficult. Aggressive single-reward optimization also risks reward hacking, where the model improves the targeted metric at the cost of diversity or other quality dimensions~\citep{flowgrpo_liu_2025}.

\begin{figure*}[t!]
\centering
\includegraphics[width=\textwidth]{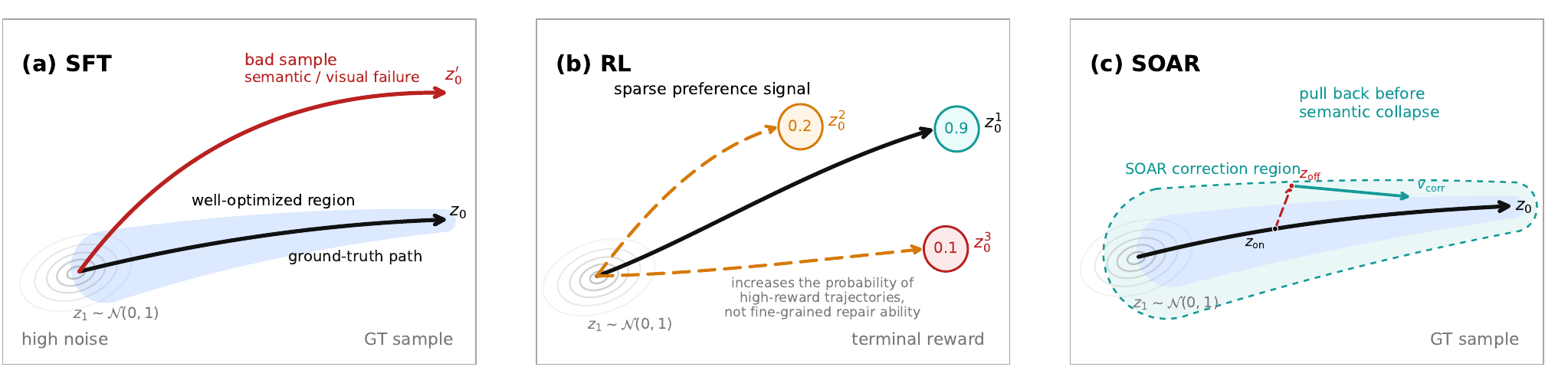}
\caption{Conceptual comparison of post-training regimes from a trajectory perspective. \textbf{Left: SFT} trains only on ground-truth denoising states. Once an inference rollout exits the narrow corridor of well-optimized states, subsequent steps sharpen the wrong structure. \textbf{Middle: RL} separates good and bad final outcomes with a terminal reward, but the sparse signal arrives after the erroneous rollout has already formed. \textbf{Right: SOAR} performs a one-step rollout to generate off-trajectory states and supervises the model to correct back toward the clean target, acting before semantic errors fully materialize.}
\label{fig:sft_rl_soar_compare}
\end{figure*}

This paper proposes \methodnamebf (Self-Correction for Optimal Alignment and Refinement), a post-training method that replaces SFT as a stronger first stage directly following pretraining, while remaining fully compatible with subsequent RL alignment. The core idea is simple: instead of waiting for a terminal reward after a full rollout, we directly teach the model how to correct its own trajectory errors at the timestep where they occur. Starting from a real training pair, \methodname performs a single stop-gradient ODE step with the current model to generate an off-trajectory state, re-noises this state to multiple auxiliary noise levels, and supervises the denoiser to steer each auxiliary state back toward the original clean target. The resulting training signal is \emph{on-policy}, since the off-trajectory states evolve with the model's own parameters; \emph{dense}, as the correction target is derived from the ground-truth sample itself rather than collapsed into a scalar reward or a relative preference; and \emph{reward-free}, requiring no external reward model or preference annotation. The correction target is derived analytically from the training pair, eliminating the credit-assignment problem entirely.

Placing \methodname in the landscape of existing post-training paradigms clarifies its role. SFT trains exclusively on ground-truth denoising states; it is off-policy by construction because the model never sees its own inference-time errors during training. RL-based methods introduce on-policy or reward-driven signals but share common limitations: the supervision is sparse and terminal, credit must be assigned across dozens of denoising steps, and aggressive single-reward optimization risks reward hacking. By design, \methodname is free from all of these limitations. Moreover, when the training corpus is curated according to a target preference, such as filtering for high aesthetic or high text-image alignment, \methodname can absorb that preference while preserving quality on other dimensions, since its dense ground-truth supervision leaves no room for reward hacking. In the current work we position \methodname as a first-stage replacement for SFT that seamlessly connects with subsequent RL alignment; however, our experiments on preference-curated subsets already demonstrate its potential as a data-efficient alternative to RL-based optimization itself. We summarize the key differences among these paradigms in Table~\ref{tab:paradigm_compare}.

\Needspace{12\baselineskip}
\noindent Our main contributions are:
\begin{itemize}[leftmargin=2em,itemsep=1pt,topsep=1pt,parsep=0pt,partopsep=0pt]
\item We identify exposure bias, the distributional mismatch between ground-truth training states and model-induced inference states, as a unifying explanation for many persistent diffusion generation failures.
\item We propose \methodname, a practical post-training algorithm that constructs off-trajectory states via a single ODE rollout, re-noises them, and optimizes the model with an analytically derived correction target anchored to the original clean sample.
\item We provide a controlled experimental comparison showing that \methodname improves both rule-based (GenEval, OCR) and model-based (PickScore, ClipScore, HPSv2.1, Aesthetic, ImageReward) metrics simultaneously over SFT, achieving monotonic reward improvement without a reward model.
\item We show that on curated high-reward data subsets, \methodname outperforms both vanilla SFT and explicit RL (Flow-GRPO) in final metric value, validating the trajectory-correction paradigm as a data-efficient alternative to reward optimization.
\end{itemize}

\begin{table*}[t]
  \centering
  \small
  \setlength{\tabcolsep}{4pt}
  \renewcommand{\arraystretch}{1.15}
  \caption{Qualitative comparison of post-training paradigms for diffusion models.
    \methodname combines on-policy training with dense, reward-free supervision
    while maintaining low training cost.}
  \label{tab:paradigm_compare}
  \resizebox{\textwidth}{!}{%
  \begin{tabular}{l cccccccc}
    \toprule
    & \makecell{Training\\Signal}
    & On-policy
    & \makecell{Signal\\Density}
    & \makecell{Reward\\Model}
    & \makecell{Credit\\Assign.}
    & \makecell{Reward\\Hacking}
    & \makecell{Resolves\\Model Bias}
    & \makecell{Training\\Cost} \\
    \midrule
    \textbf{SFT}  & Supervised (GT traj.)              & \xmark & Dense  & \xmark  & \xmark  & \xmark  & \xmark  & Low    \\
    \textbf{DPO}  & Pref.\ pairs (implicit reward)     & \xmark & Sparse & \xmark  & \xmark  & \xmark  & Partial & Medium \\
    \textbf{RL}   & Reward function                     & \cmark & Sparse & \cmark  & \cmark  & \cmark  & Partial & High   \\
    \hdashline\noalign{\vskip 2pt}
    \textbf{SOAR} & Supervised (traj.\ correction)      & \cmark~(single-step) & Dense  & \xmark  & \xmark  & \xmark  & \cmark  & Low    \\
    \bottomrule
  \end{tabular}}
\end{table*}

%% file: sec/2_methodology.tex
\section{Methodology}
\label{sec:method}

\subsection{Preliminaries: Flow Matching}
\label{sec:prelim}

Let $z_0 \sim q(z_0)$ denote a clean latent sampled from the data distribution and $z_1 \sim \mathcal{N}(0, I)$ denote Gaussian noise. Rectified flow~\citep{liu_rectifiedflow_2023,sd35_esser_2024} defines a linear interpolation path between data and noise:
\begin{equation}
z_t = (1-t)\,z_0 + t\,z_1, \qquad t \in [0,1],
\label{eq:interpolation}
\end{equation}
where we use $t$ (or equivalently the scheduler value $\sigma_t$) to index the noise level. A neural network $v_\theta(z_t, c, t)$ is trained to predict the velocity field by minimizing the flow matching objective:
\begin{equation}
\mathcal{L}_{\mathrm{FM}}(\theta) = \mathbb{E}_{z_0, z_1, c, t}\bigl[\lVert v_\theta(z_t, c, t) - (z_1 - z_0) \rVert^2\bigr].
\label{eq:fm_loss}
\end{equation}
Given a trained velocity field, sampling proceeds by solving the ODE from $t{=}1$ (pure noise) to $t{=}0$ (clean data). With Euler discretization over $K$ steps, each update reads
\begin{equation}
z_{t-\Delta t} = z_t + \Delta t \cdot v_\theta(z_t, c, t), \qquad \Delta t = -1/K.
\end{equation}
The velocity prediction at any state $z$ and noise level $\sigma$ implies a \emph{clean-endpoint estimate}:
\begin{equation}
\hat{z}_0(z, \sigma) = z - \sigma \cdot v_\theta(z, c, \sigma).
\label{eq:clean_endpoint}
\end{equation}
This map will play a central role in the correction objective below.

\subsection{Post-Training Paradigms}
\label{sec:paradigms}

\paragraph{Supervised fine-tuning (SFT).}
Given a curated dataset of $(z_0, c)$ pairs, SFT continues to minimize the flow matching loss in Eq.~\eqref{eq:fm_loss}. The training states $z_t$ are constructed by the forward process of real samples: $z_t = (1{-}t)z_0 + tz_1$ with fresh noise $z_1 \sim \mathcal{N}(0,I)$. This means SFT only optimizes the model on the \emph{ground-truth} state distribution
\begin{equation}
p_t^{\mathrm{data}}(z) = \int \delta\bigl((1{-}t)z_0 + tz_1\bigr)(z)\, q(z_0)\,\mathcal{N}(z_1)\,dz_0\,dz_1.
\label{eq:p_data}
\end{equation}
During inference, however, the model generates states from a different distribution $p_t^\theta(z)$, defined implicitly by the ODE rollout from $z_1$ through the learned velocity field. Since each state depends on the model's previous predictions, early errors propagate and push later states into regions where the model was never directly trained. This train--inference mismatch is the \emph{exposure bias} of diffusion models: the same phenomenon that affects autoregressive generation~\citep{bengio_scheduled_2015}, but accumulated along the denoising trajectory instead of the token sequence.

\paragraph{Reward-based alignment and RL.}
To improve beyond SFT, recent work explores reward-driven optimization for diffusion and flow-matching models. Approaches range from offline preference learning such as DPO~\citep{wallace_diffusiondpo_2024}, which adjusts the relative likelihood of preferred and dispreferred samples, to online policy-gradient methods. Among the latter, Flow-GRPO~\citep{flowgrpo_liu_2025} is representative: it converts the deterministic ODE sampler into an equivalent SDE that preserves the marginal distribution at all timesteps, enabling stochastic exploration, and applies group-relative policy optimization (GRPO) where a group of $G$ images is generated per prompt, scored by a reward function $r(z_0, c)$, and the normalized group advantage $\hat{A}^i = (r^i - \bar{r})/\mathrm{std}(r)$ drives a clipped policy-gradient update. While Flow-GRPO achieves strong results on compositional benchmarks (GenEval 0.63$\to$0.95), RL-based approaches share inherent limitations: (i)~the reward signal is sparse and terminal, arriving only after the full denoising trajectory is complete; (ii)~credit assignment across dozens of denoising steps is difficult; (iii)~aggressive single-reward optimization risks reward hacking, where the targeted metric improves at the cost of diversity or other quality dimensions.

\subsection{SOAR: Trajectory Bias Correction}
\label{sec:soar}

\methodname addresses the exposure bias directly: instead of patching failures after a full rollout, it teaches the model to correct trajectory errors at the timestep where they occur.

\begin{figure}[t]
\centering
\includegraphics[width=0.8\linewidth]{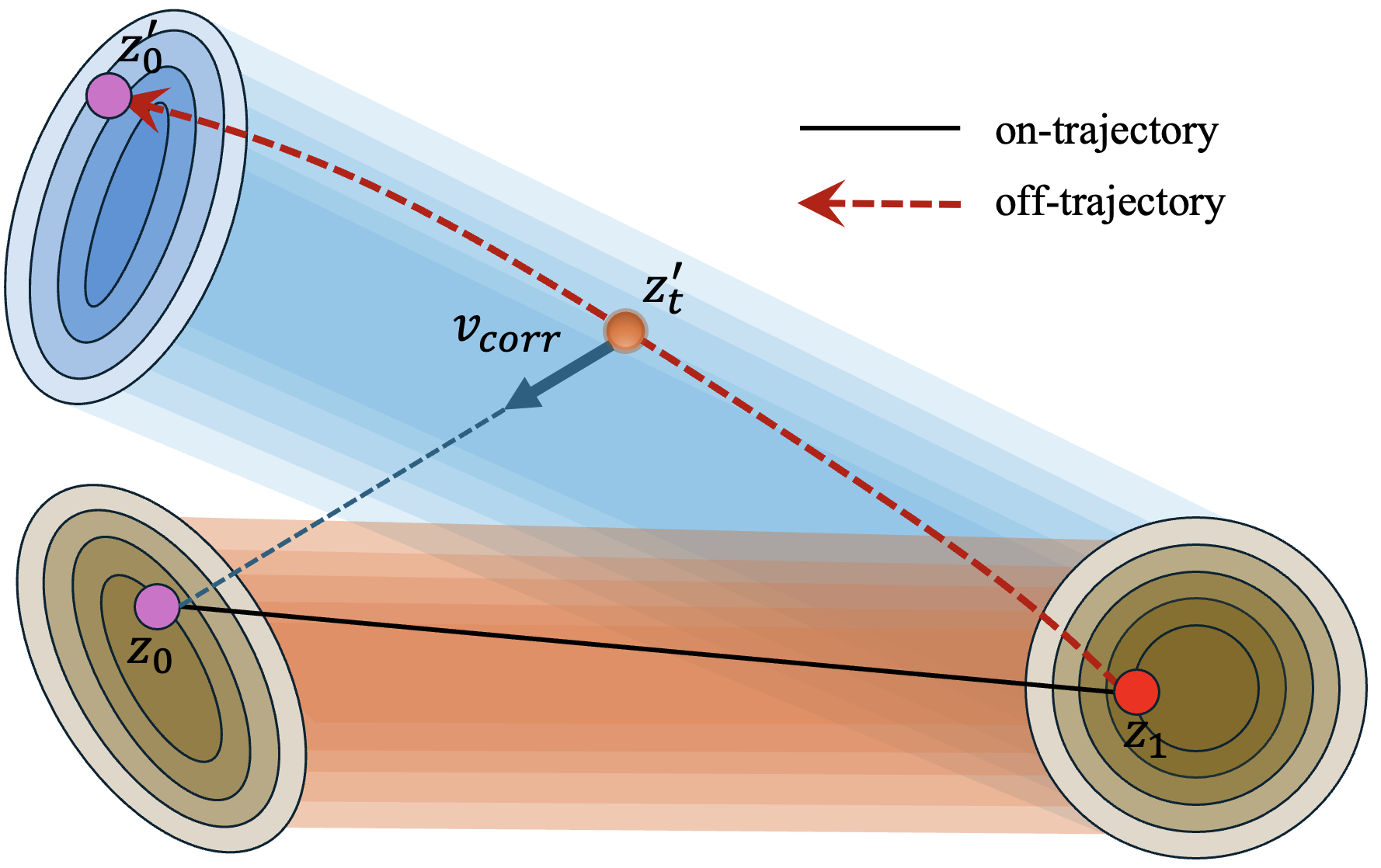}
\caption{Noise-level analysis illustrating the ideal trajectory, the biased rollout, and the local correction direction. The off-trajectory state is constructed by one ODE step with the current model and then re-noised to auxiliary levels.}
\label{fig:noise_diagnosis}
\end{figure}

\subsubsection{Distributional Mismatch}
\label{sec:mismatch}

SFT minimizes the velocity error over ground-truth states $z \sim p_t^{\mathrm{data}}$ (Eq.~\ref{eq:p_data}). At inference time the model instead encounters states $z \sim p_t^\theta$, the distribution induced by its own ODE rollout. Because each sampled state is produced from the model's previous prediction, even small local velocity errors can move the trajectory away from the ground-truth path, and the resulting deviation compounds across subsequent denoising steps. \methodname closes this gap by explicitly constructing states that approximate $p_t^\theta$ during training and supervising the model on them.

\subsubsection{Constructing Off-Trajectory States}
\label{sec:off_traj}

Given a training pair $(z_0, c)$ and sampled noise $z_1 \sim \mathcal{N}(0,I)$, we construct the on-trajectory state at a sampled time $t_0$:
\begin{equation}
z_{\sigma_{t_0}} = (1-\sigma_{t_0})\,z_0 + \sigma_{t_0}\,z_1.
\end{equation}
We then simulate what the model would produce during inference by running a single \emph{stop-gradient} ODE step with classifier-free guidance (CFG). Concretely, we compute a detached CFG velocity $v_{\mathrm{cfg}} = \operatorname{sg}[v_{\mathrm{uncond}} + w_{\mathrm{cfg}}(v_{\mathrm{cond}} - v_{\mathrm{uncond}})]$ at the shared state $z_{\sigma_{t_0}}$, and define the next timestep $t_1 = \max(t_0 - 1/K,\, 0)$. One Euler step produces the off-trajectory state:
\begin{equation}
\hat{z}_{\sigma_{t_1}} = z_{\sigma_{t_0}} + (\sigma_{t_1} - \sigma_{t_0})\,v_{\mathrm{cfg}}.
\label{eq:ode_step}
\end{equation}
This is the model's own prediction; it deviates from the ideal state $(1{-}\sigma_{t_1})z_0 + \sigma_{t_1}z_1$ in exactly the way that inference would.

To generate diverse auxiliary supervision, we \emph{re-noise} this off-trajectory state by interpolating it back toward the noise endpoint $z_1$. For $N$ uniformly sampled auxiliary noise levels $\sigma_{t'} \sim \mathrm{Uniform}[\sigma_{t_1}, 1]$, we set $\alpha = (\sigma_{t'} - \sigma_{t_1})/(1 - \sigma_{t_1})$ and construct:
\begin{equation}
z_{\sigma_{t'}} = (1-\alpha)\,\hat{z}_{\sigma_{t_1}} + \alpha\, z_1.
\label{eq:renoise}
\end{equation}

A key design choice is that re-noising uses the \textbf{same} $z_1$ as the base loss. This keeps the auxiliary state in the neighborhood of the original transport ray $z_0 \leftrightarrow z_1$: the deviation from the ideal path is $(1{-}\alpha)$ times the one-step rollout error, which is bounded and shrinks as $\alpha \to 1$. Consequently, $z_0$ remains the uniquely correct clean-endpoint target for these nearby off-trajectory states, and the correction direction has a clear geometric meaning along the conditional interpolation path (see Appendix~\ref{app:diffusion_derivation} for the formal argument).

\subsubsection{Correction Objective}
\label{sec:correction}

The most natural way to correct a drifted trajectory is \emph{local trajectory matching}: require that, after a small step $\Delta t$, the off-trajectory state reconverges with the on-trajectory state:
\begin{equation}
z_\sigma + \Delta t \cdot v_{\mathrm{gt}} = z_{\sigma}' + \Delta t \cdot v_{\mathrm{corr}},
\label{eq:local_match}
\end{equation}
where $z_\sigma$ is the on-trajectory state and $z_{\sigma}'$ is the off-trajectory state at the same noise level. This ensures the corrected path returns to the ideal trajectory after one step. However, the step size $\Delta t$ introduces an ambiguity: different $\Delta t$ yield different targets, and the optimal choice depends on the downstream trajectory, making it impractical in a multi-step ODE.

We resolve this by requiring both states to reach the same \emph{final} endpoint $z_0$. On the ideal trajectory, the on-trajectory state $z_\sigma = (1{-}\sigma)z_0 + \sigma z_1$ with velocity $v_{\mathrm{gt}} = z_1 - z_0$ satisfies
\begin{equation}
z_\sigma - \sigma \cdot v_{\mathrm{gt}} = z_0.
\label{eq:on_goal}
\end{equation}
We impose the same \emph{goal consistency} condition on the off-trajectory state. Since $z_{\sigma}'$ remains close to the original trajectory by construction (Section~\ref{sec:off_traj}), the clean target $z_0$ is still the correct destination:
\begin{equation}
z_{\sigma}' - \sigma \cdot v_{\mathrm{corr}} = z_0.
\label{eq:off_goal}
\end{equation}
Solving for $v_{\mathrm{corr}}$ yields the closed-form correction target:
\begin{equation}
v_{\mathrm{corr}} = \frac{z_{\sigma}' - z_0}{\sigma}.
\label{eq:vcorr}
\end{equation}
This eliminates the $\Delta t$ ambiguity: the target depends only on the current state and the clean anchor $z_0$. When $z_{\sigma}' = z_\sigma$ (no deviation), it reduces exactly to $v_{\mathrm{gt}} = z_1 - z_0$, recovering the standard flow matching target. When $z_{\sigma}' \neq z_\sigma$, $v_{\mathrm{corr}} - v_{\mathrm{gt}} = (z_{\sigma}' - z_\sigma)/\sigma$ is a correction proportional to the rollout deviation, steering the model back toward $z_0$ from the off-trajectory states it actually visits during inference. This is the dense, per-timestep signal that SFT cannot provide.

The total \methodname objective combines the on-trajectory base loss with the auxiliary correction loss. Concretely, for a batch of $B$ samples with $N$ auxiliary points each:
\begin{align}
\mathcal{L}_{\mathrm{base}} &= \sum_{b=1}^{B} w(\sigma_{t_0}^{(b)})\,
  \lVert v_\theta(z_{\sigma_{t_0}}^{(b)}, c, t_0) - v_{\mathrm{gt}}^{(b)}\rVert^2,
  \label{eq:loss_base} \\
\mathcal{L}_{\mathrm{corr}} &= \sum_{p=1}^{P} w(\sigma_{t'}^{(p)})\,
  \lVert v_\theta(z_{\sigma_{t'}}^{(p)}, c, t') - v_{\mathrm{corr}}^{(p)}\rVert^2,
  \label{eq:loss_corr}
\end{align}
where $P$ is the total number of valid auxiliary points. The combined objective is normalized by the total count of supervised items (synchronized across distributed workers):
\begin{equation}
\mathcal{L}_{\mathrm{SOAR}} = \frac{\mathcal{L}_{\mathrm{base}} + \lambda\,\mathcal{L}_{\mathrm{corr}}}{B + \lambda P}.
\label{eq:soar_loss}
\end{equation}
From a distribution-matching perspective, both terms minimize the 2-Wasserstein distance between the model's clean-endpoint prediction and the true $z_0$, but the expectation is now taken over $p^{\mathrm{data}} \cup p^{\mathrm{rollout}}$ instead of $p^{\mathrm{data}}$ alone, directly closing the exposure-bias gap (Appendix~\ref{app:dist_matching}).

\begin{algorithm}[t]
\caption{\methodname training (ODE-only, default configuration)}
\label{alg:soar}
\begin{algorithmic}[1]
\Require velocity field $v_\theta$, CFG scale $w_{\mathrm{cfg}}$, step count $K$, aux count $N$, weight $\lambda$
\For{each training batch $(z_0, c)$}
    \State Sample $z_1 \sim \mathcal{N}(0,I)$, sample time $t_0$, compute $\sigma_{t_0}$
    \State $z_{\sigma_{t_0}} \gets (1-\sigma_{t_0})z_0 + \sigma_{t_0} z_1$ \Comment{on-trajectory state}
    \State $v_{\mathrm{on}} \gets v_\theta(z_{\sigma_{t_0}}, c, t_0)$;\quad $v_{\mathrm{gt}} \gets z_1 - z_0$
    \State Accumulate $\mathcal{L}_{\mathrm{base}} \gets w(\sigma_{t_0})\lVert v_{\mathrm{on}} - v_{\mathrm{gt}}\rVert^2$
    \State $v_{\mathrm{cfg}} \gets \operatorname{sg}[v_{\mathrm{uncond}} + w_{\mathrm{cfg}}(v_{\mathrm{cond}} - v_{\mathrm{uncond}})]$
    \State $t_1 \gets \max(t_0 - 1/K,\, 0)$;\quad $\sigma_{t_1} \gets \sigma(t_1)$
    \State $\hat{z}_{\sigma_{t_1}} \gets z_{\sigma_{t_0}} + (\sigma_{t_1} - \sigma_{t_0})\,v_{\mathrm{cfg}}$ \Comment{one ODE step}
    \For{$n = 1, \dots, N$}
        \State Sample $\sigma_{t'} \sim \mathrm{Uniform}[\sigma_{t_1}, 1]$
        \State $\alpha \gets (\sigma_{t'} - \sigma_{t_1})/(1 - \sigma_{t_1})$
        \State $z_{\sigma_{t'}} \gets (1{-}\alpha)\,\hat{z}_{\sigma_{t_1}} + \alpha\, z_1$ \Comment{re-noise with same $z_1$}
        \State $v_{\mathrm{off}} \gets v_\theta(z_{\sigma_{t'}}, c, t')$
        \State $v_{\mathrm{corr}} \gets (z_{\sigma_{t'}} - z_0) / \sigma_{t'}$
        \State Accumulate $\mathcal{L}_{\mathrm{corr}} \mathrel{+}= w(\sigma_{t'})\lVert v_{\mathrm{off}} - v_{\mathrm{corr}}\rVert^2$
    \EndFor
    \State $\mathcal{L}_{\mathrm{SOAR}} \gets (\mathcal{L}_{\mathrm{base}} + \lambda\,\mathcal{L}_{\mathrm{corr}}) / (1 + \lambda N)$ \Comment{normalized}
    \State Update $\theta$ using $\nabla_\theta \mathcal{L}_{\mathrm{SOAR}}$
\EndFor
\end{algorithmic}
\end{algorithm}

Unlike reward-based post-training, \methodname does not wait until a full sample is generated and then ask an external reward model to score the outcome. It operates directly at the denoising level and corrects trajectory bias where the failure first begins to form. A detailed comparison with SFT, DPO, and Flow-GRPO is provided in the Introduction; the full derivation of the correction target and the distribution-matching interpretation are given in Appendix~\ref{app:diffusion_derivation}.

%% file: sec/3_experiment.tex
\section{Experiments}
\label{sec:experiments}

\subsection{Experimental Setup}
\label{sec:exp_setup}

\paragraph{Backbone and evaluation.}
All experiments use the \basemodel~\citep{sd35_esser_2024} backbone. Generated images are evaluated at $512\times512$ resolution with $\mathrm{cfg}{=}4.5$. Our model-based metrics are evaluated on out-of-domain (OOD) DrawBench~\citep{imagen_saharia_2022}. For rule-based metrics, GenEval~\citep{t2icompbench_huang_2023} (compositional accuracy) and OCR (text rendering fidelity) are evaluated on the corresponding GenEval and OCR test sets from Flow-GRPO~\citep{flowgrpo_liu_2025}, following the evaluation protocol used by DiffusionNFT~\citep{zheng_diffusionnft_2026}. These test sets are used for evaluation only, not for training.

\paragraph{Training data.}
\methodname is trained on paired image--text data and requires no preference labels, reward annotations, or negative samples. The full training corpus contains \textbf{286,119} image--caption pairs. Because the correction target is anchored to the clean latent of a real sample, data quality is critical: noisy captions or semantically mismatched pairs would inject incorrect anchors into the supervision signal.

\paragraph{Broad-data comparison.}
Our primary experiment trains all methods on the \textbf{full 286K corpus} and compares them head-to-head across all metrics. The comparison set includes the SD3.5-M pretrained baseline (no post-training), +\,SFT (full-parameter fine-tuning, 10k steps), and +\,\methodname (full-parameter trajectory-corrected fine-tuning, 10k steps). For context, we also report reference baselines from SD-XL~\citep{sdxl_podell_2024}, SD3.5-L~\citep{sd35_esser_2024}, and FLUX.1-Dev~\citep{Flux_blacklabs_2025}. All runs use a global batch size of 512 and a constant learning rate of $2\times10^{-5}$. This setting directly measures whether trajectory correction improves general generation quality beyond what standard SFT achieves.

\paragraph{Reward-specific comparison.}
To isolate the effect of the training \emph{objective} rather than data scale, we construct two curated subsets from the full corpus: a \textbf{High-Aesthetic} subset (aesthetic score $\ge 6.8$, 3{,}725 pairs) and a \textbf{High-ClipScore} subset (CLIP score $\ge 0.40$, 6{,}857 pairs). On each subset we compare three paradigms under identical compute (5k gradient steps, 128 GPUs, global batch size 32, constant learning rate $2\times10^{-5}$): SFT and \methodname both perform full-parameter fine-tuning with only an implicit reward signal from the curated data, while Flow-GRPO~\citep{flowgrpo_liu_2025} explicitly optimizes the target reward via LoRA-based policy gradients with direct access to the reward model. Because all methods see the same high-quality data, any performance difference must come from the optimization paradigm itself, revealing which approach is more data-efficient and which is more prone to representation drift or reward hacking.

\subsection{Main Results}
\label{sec:main_results}

\begin{table}[t]
  \centering
  \small
  \setlength{\tabcolsep}{4.5pt}
  \renewcommand{\arraystretch}{1.12}
  \caption{Main quantitative comparison at $512{\times}512$ ($\mathrm{cfg}{=}4.5$). Rule-based metrics are evaluated on the Flow-GRPO test sets; model-based metrics are evaluated on OOD DrawBench. $^\dagger$Official checkpoint. $^\ddagger$At $1024{\times}1024$. \textbf{Bold}: best overall; \underline{underline}: second best.}
  \label{tab:sd35m_512_main}
  \resizebox{0.98\linewidth}{!}{%
  \begin{tabular}{l c cc ccccc}
    \toprule
    \multirow{2}{*}{Model} & \multirow{2}{*}{\#Iter}
    & \multicolumn{2}{c}{\shortstack{Rule-Based\\(Flow-GRPO test sets)}}
    & \multicolumn{5}{c}{\shortstack{Model-Based\\(OOD DrawBench)}} \\
    \cmidrule(lr){3-4}
    \cmidrule(lr){5-9}
    & & GenEval & OCR & PickScore & ClipScore & HPSv2.1 & Aesthetic & ImgRwd \\
    \midrule
    SD-XL$^\ddagger$          & -- & 0.55 & 0.14 & 22.42 & 0.287 & 0.280 & 5.60 & 0.76 \\
    SD3.5-L$^\ddagger$        & -- & {0.71} & {0.68} & {22.91} & 0.289 & {0.288} & 5.50 & 0.96 \\
    FLUX.1-Dev$^\dagger$      & -- & 0.66 & 0.59 & 22.84 & {0.295} & 0.274 & {5.71} & 0.96 \\
    \midrule
    SD3.5-M                   & -- & 0.63 & 0.59 & 22.34 & 0.285 & 0.279 & 5.36 & 0.85 \\
    \quad + SFT               & 10k & 0.70 & 0.64 & 22.71 & {0.295} & 0.284 & 5.35 & {1.04} \\
    \quad + \textbf{SOAR (Ours)} & 10k & \textbf{0.78} & \textbf{0.67} & \textbf{22.86} & \textbf{0.295} & \textbf{0.289} & \textbf{5.46} & \textbf{1.09} \\
    \bottomrule
  \end{tabular}%
  }
\end{table}

Table~\ref{tab:sd35m_512_main} summarizes the broad-data comparison results. \methodname improves over SFT on \emph{every} reported metric: on the Flow-GRPO test sets, GenEval rises from 0.70 to 0.78 (+11\% relative) and OCR from 0.64 to 0.67; on OOD DrawBench, all five model-based scores improve simultaneously. The rule-based gains are particularly relevant because GenEval and OCR directly measure compositional accuracy and text-rendering fidelity, precisely the failure modes tied to early semantic decisions in the denoising trajectory, suggesting that trajectory-level correction prevents errors at high-noise stages from propagating into visible failures. Notably, the OOD DrawBench model-based improvements (PickScore +0.15, HPSv2.1 +0.005, Aesthetic +0.11, ImageReward +0.05 over SFT) are achieved \emph{without} any reward model during training. On-trajectory denoising preserves the coherent global structure established at high-noise stages, and preference models implicitly reward this structural consistency when scoring the final image. Compared with reference models, \methodname on SD3.5-M surpasses the larger SD3.5-L on Flow-GRPO test-set GenEval (0.78 vs.\ 0.71) and OOD DrawBench HPSv2.1 (0.289 vs.\ 0.288), and approaches FLUX.1-Dev on OOD DrawBench PickScore (22.86 vs.\ 22.84), demonstrating that trajectory correction can close much of the gap between a medium-sized model and significantly larger architectures.

\subsection{Reward-Specific Training Dynamics}
\label{sec:reward_dynamics}

\begin{figure*}[t]
\centering
\includegraphics[width=\textwidth]{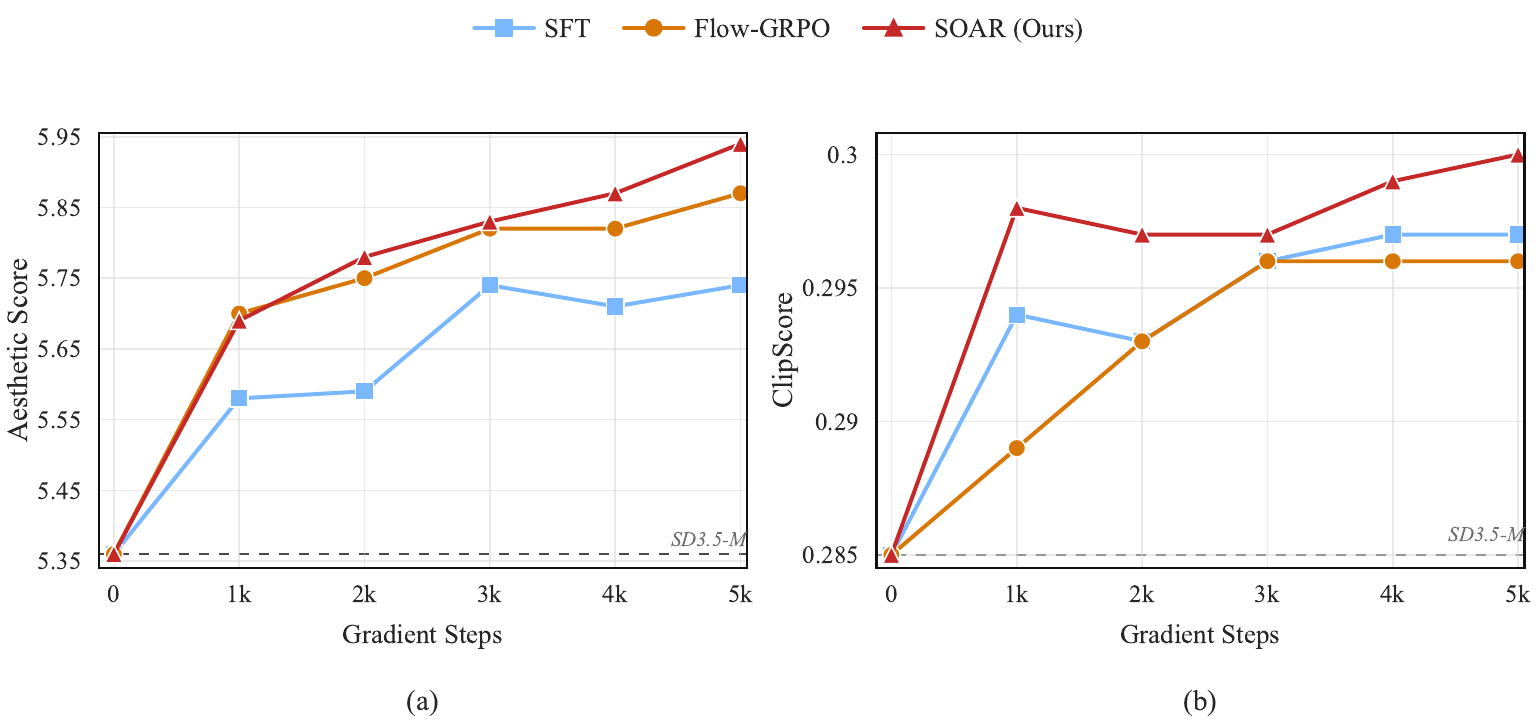}
\caption{\small Reward-specific training dynamics at $512{\times}512$. \textbf{(a)}~OOD DrawBench Aesthetic score after training on the 3{,}725-pair High-Aesthetic subset ($\ge6.8$). \textbf{(b)}~OOD DrawBench ClipScore after training on the 6{,}857-pair High-ClipScore subset. All methods start from the same SD3.5-M pretrained checkpoint (step\,0). SOAR achieves the highest final values in both settings with monotonic improvement.}
\label{fig:reward_curves}
\end{figure*}

We now turn to the reward-specific setting described in Section~\ref{sec:exp_setup}: SFT, \methodname, and Flow-GRPO are trained on the High-Aesthetic and High-ClipScore subsets under identical compute, contrasting implicit data-driven optimization against explicit RL. These subsets are used for training; Figure~\ref{fig:reward_curves} reports the corresponding target metrics on the OOD DrawBench evaluation prompts across training steps.

\paragraph{SOAR achieves the highest reward with monotonic improvement.}
On OOD DrawBench, \methodname reaches the best final value at 5{,}000 steps in both reward-specific experiments, reaching 5.94 aesthetic score (vs.\ 5.74 for SFT, 5.87 for GRPO) and 0.300 ClipScore (vs.\ 0.297 for SFT, 0.296 for GRPO). Its improvement curve is \emph{monotonic}: every checkpoint improves over the last, because the auxiliary correction loss regularizes the denoising trajectory and prevents representation drift. In contrast, SFT saturates early, jumping from the 5.36 baseline to 5.58 aesthetic at 1k steps but plateauing at 5.74 for the remaining 4k steps (ClipScore similarly plateaus at 0.297). This is consistent with the exposure-bias analysis: standard fine-tuning quickly absorbs the dominant data distribution at on-trajectory states but lacks a mechanism to further improve off-trajectory behavior, so additional training steps yield diminishing returns.

\paragraph{Explicit RL does not dominate implicit approaches.}
Perhaps surprisingly, Flow-GRPO, which has direct access to the reward model, does not reach the highest final value in either experiment. On aesthetics it achieves 5.87 (vs.\ \methodname's 5.94); on ClipScore it reaches 0.296 (vs.\ \methodname's 0.300). Two factors explain this: (i)~GRPO updates only LoRA parameters ($\sim$75\,MB vs.\ full model), limiting representational capacity; (ii)~explicit reward optimization through RL is sample-inefficient compared with direct supervision on high-quality data; each RL gradient step requires generating multiple rollouts and computing rewards, whereas \methodname derives a dense correction signal analytically from each training pair.

\paragraph{GRPO trades off non-target metrics.}
Although GRPO improves the target aesthetic score, its ClipScore drops from 0.285 (baseline) to 0.277 at 5{,}000 steps, a 2.8\% relative decline. In contrast, SFT and \methodname maintain or improve ClipScore while boosting aesthetics. This illustrates the well-known reward-hacking risk: aggressive single-reward optimization can degrade uncorrelated properties. \methodname avoids this pitfall entirely because it has no reward function to overfit to; its implicit ``reward'' is the data distribution itself, which naturally balances multiple quality dimensions.

These results motivate the staged pipeline proposed in this paper: first use \methodname to correct trajectory-level bias on broad data (improving general quality without reward hacking), then apply targeted reward optimization (e.g., GRPO) only when a specific metric deficit remains.

\subsection{Ablation Study}
\label{sec:ablation}

We ablate key design choices of \methodname. All variants are trained for 10{,}000 steps on the full training set at $512\times512$ with $\mathrm{cfg}{=}4.5$.

\begin{table}[t]
\centering
\small
\setlength{\tabcolsep}{3.8pt}
\renewcommand{\arraystretch}{1.15}
\caption{Ablation of SOAR design choices on SD3.5-M ($512\times512$, cfg$=$4.5, 10k steps). Rule-based metrics use the Flow-GRPO test sets; model-based metrics use OOD DrawBench. Best per column in \textbf{bold}.}
\label{tab:soar_ablation}
\resizebox{0.98\linewidth}{!}{%
\begin{tabular}{l cc ccccc}
\toprule
\multirow{2}{*}{Variant} &
\multicolumn{2}{c}{\shortstack{Rule-Based\\(Flow-GRPO test sets)}} &
\multicolumn{5}{c}{\shortstack{Model-Based\\(OOD DrawBench)}} \\
\cmidrule(lr){2-3} \cmidrule(lr){4-8}
& GenEval & OCR & PickScore & ClipScore & HPSv2.1 & Aesthetic & ImgRwd \\
\midrule
SOAR (ODE-only, shared $z_1$)
  & \textbf{0.78} & \textbf{0.67} & 22.86 & 0.295 & \textbf{0.289} & 5.46 & \textbf{1.09} \\
\quad $+$ SDE branch
  & 0.78 & 0.65 & 22.87 & \textbf{0.296} & 0.287 & \textbf{5.47} & 1.07 \\
\quad $+$ random $z_1$ per path
  & 0.76 & 0.64 & \textbf{22.89} & 0.292 & 0.287 & 5.45 & 1.08 \\
\bottomrule
\end{tabular}%
}
\end{table}

\paragraph{ODE-only vs.\ ODE+SDE branches.}
Beyond the deterministic ODE step described in Section~\ref{sec:off_traj}, one can construct additional stochastic rollout branches via SDE samplers (e.g., coefficient-preserving sampling). We compare the default ODE-only configuration ($M{=}1$) against 1\,ODE + 1\,SDE branch ($M{=}2$). As shown in Table~\ref{tab:soar_ablation}, removing the SDE branch slightly improves GenEval (tied at 0.78), OCR (0.67 vs.\ 0.65), HPSv2.1 (0.289 vs.\ 0.287), and ImageReward (1.09 vs.\ 1.07), while other metrics remain comparable. For the current training budget, the deterministic ODE path alone supplies sufficiently diverse off-trajectory states for bias correction. The ODE-only variant also halves the auxiliary compute cost per training step, making it the more practical default.

\paragraph{Shared $z_1$ vs.\ fresh random $z_1$.}
The default \methodname re-noises off-trajectory states using the \textbf{same} $z_1$ drawn for the base loss, keeping auxiliary states on the same interpolation path (Section~\ref{sec:off_traj}). An alternative draws independent Gaussian noise for each auxiliary path. Fresh $z_1$ slightly degrades rule-based metrics (GenEval 0.76, OCR 0.64) but yields the highest PickScore (22.89). The degradation in compositional accuracy confirms the geometric argument from Section~\ref{sec:off_traj}: reusing the original $z_1$ keeps the auxiliary states close to the actual forward-process distribution and maintains the validity of $z_0$ as the correction anchor. Random noise introduces more diverse but noisier supervision where the anchor assumption weakens, which may require longer training or careful weighting to match the default.

A full sweep of the rollout-path count $M$, auxiliary-point count $N$, step count $K$, and noise weighting $w(\sigma)$ is deferred to an extended version.

\begin{figure*}[!htb]
\centering
\includegraphics[width=\textwidth]{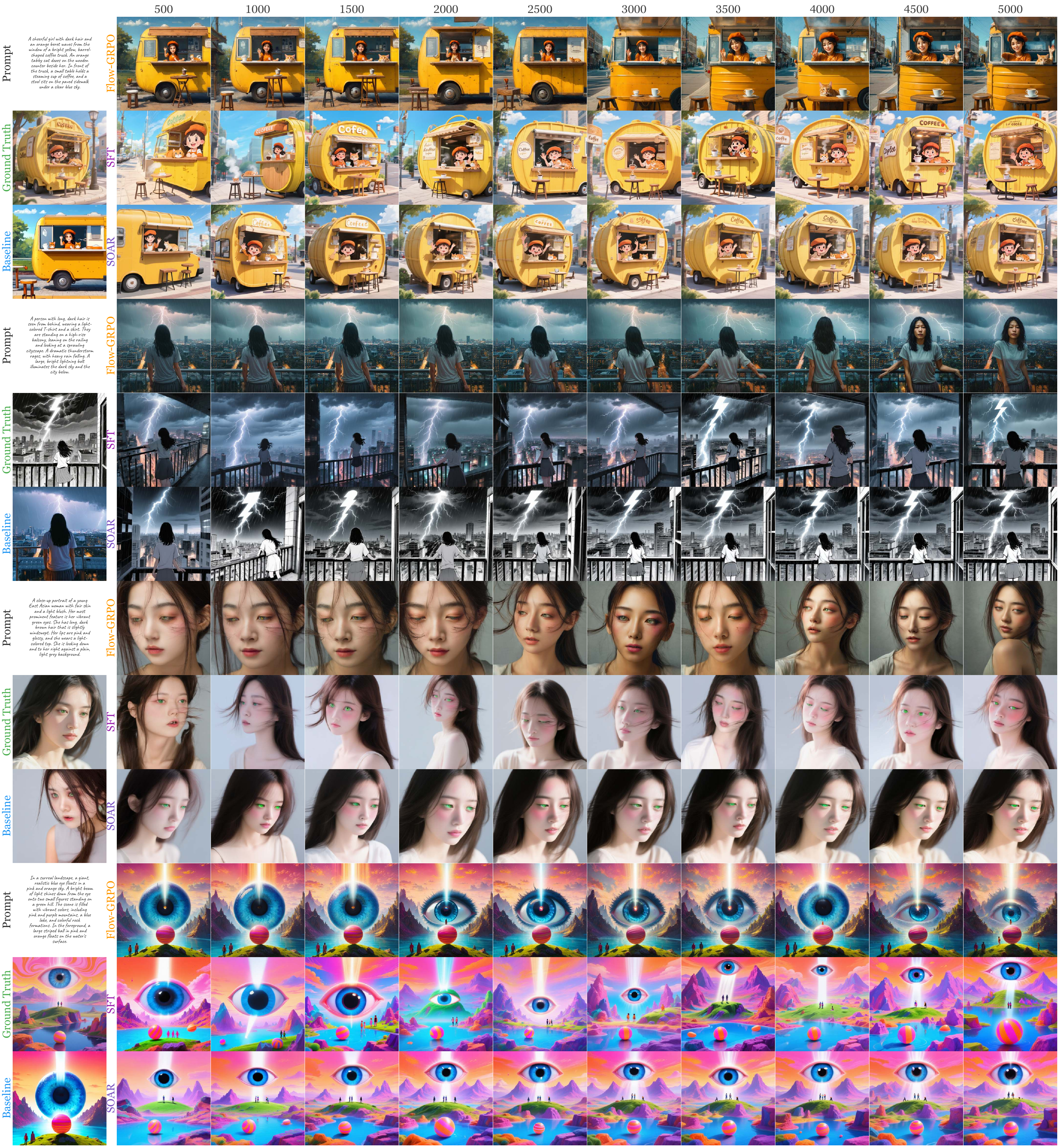}
\caption{Training visualization on the \textbf{high-aesthetic} subset, covering portraits, landscapes, and anime styles. \methodname rapidly acquires coherent composition and structural fidelity as training progresses.}
\label{fig:vis_aesth}
\end{figure*}

\begin{figure*}[p]
\centering
\includegraphics[width=\textwidth]{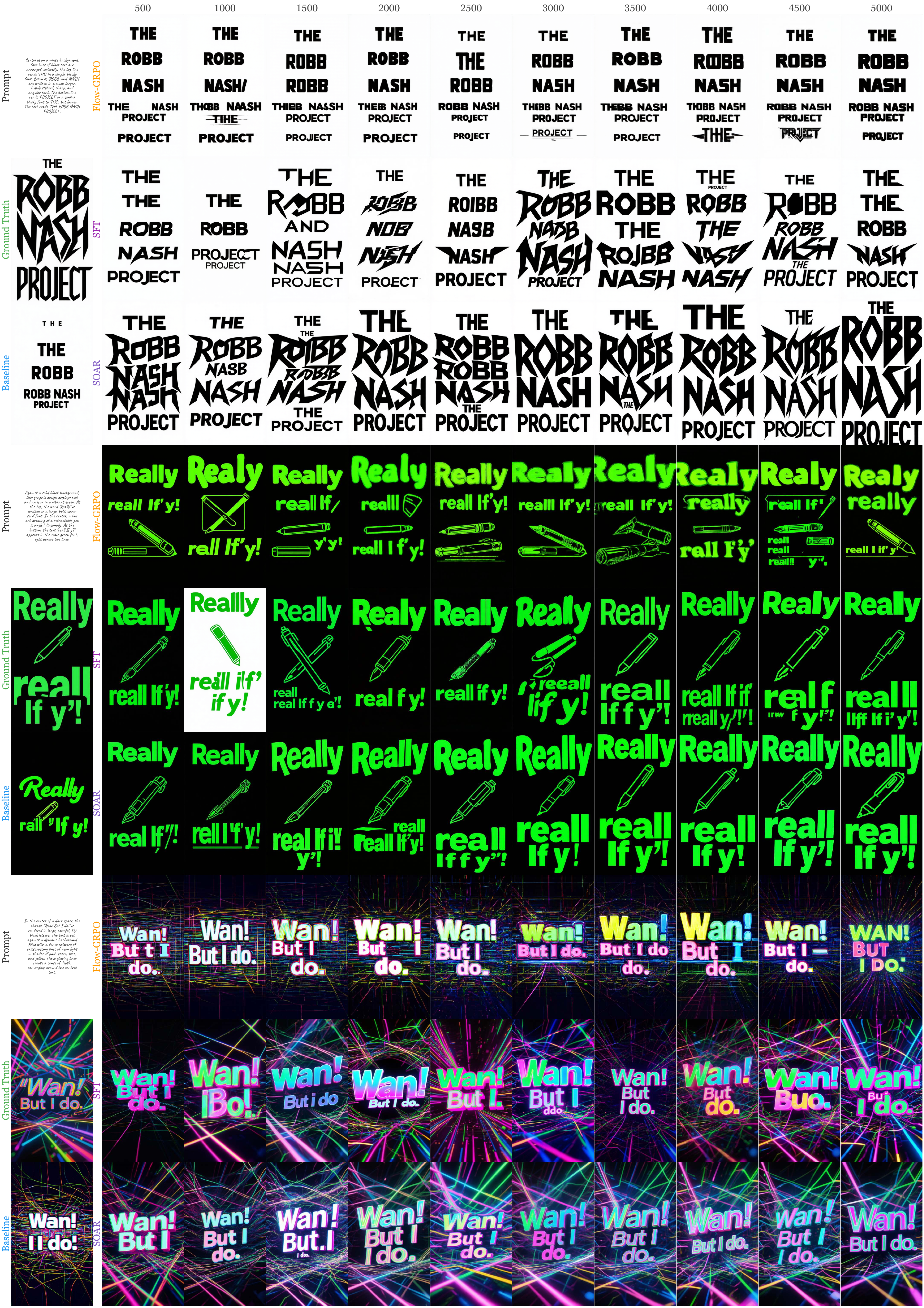}
\caption{Training visualization on \textbf{poster} data. \methodname learns layout structure, typographic placement, and visual hierarchy with high fidelity.}
\label{fig:vis_poster}
\end{figure*}

\begin{figure*}[!htb]
\centering
\includegraphics[width=\textwidth]{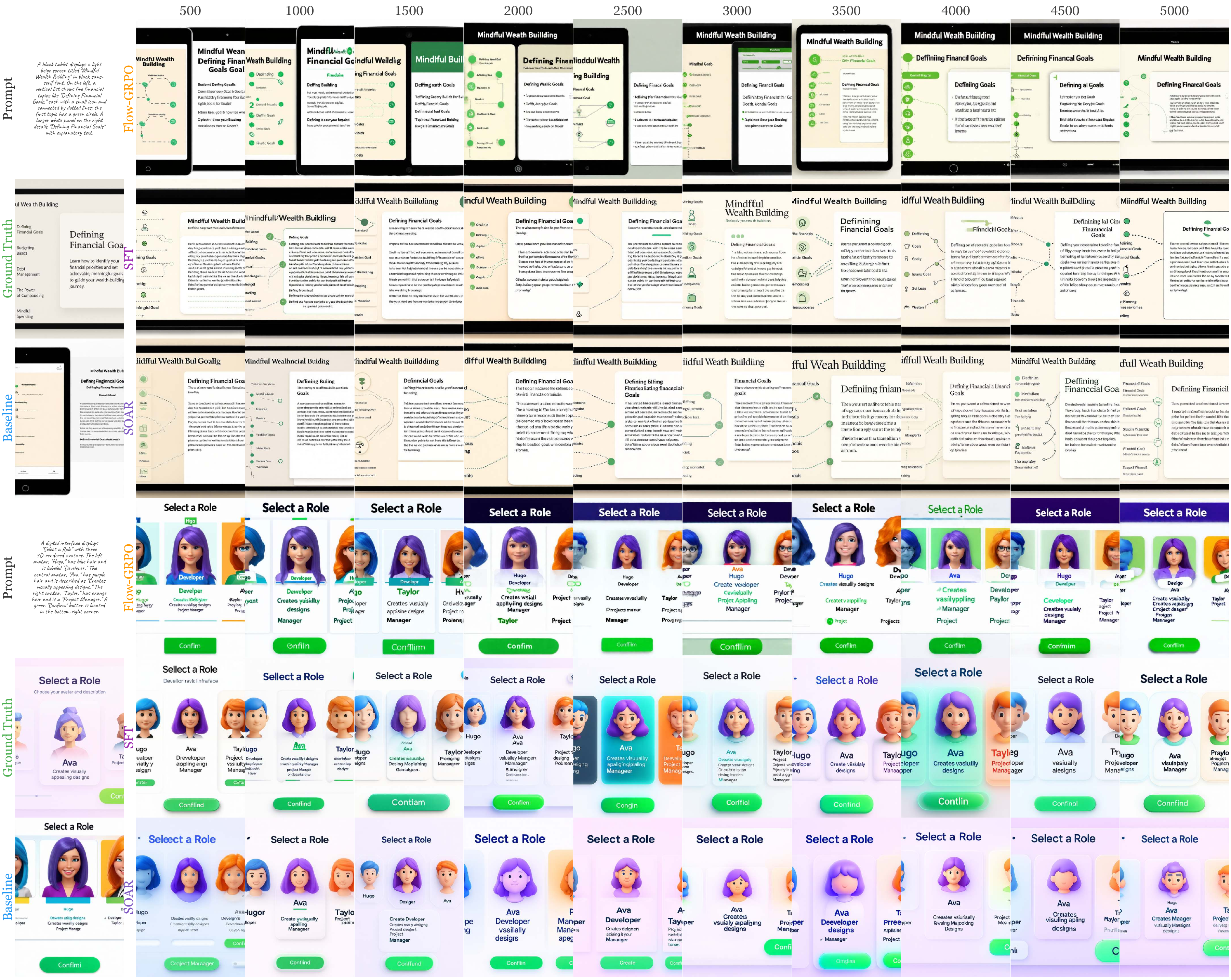}
\caption{Training visualization on \textbf{WebUI} data. \methodname accurately captures interface elements, shadow effects, and fine-grained stylistic details.}
\label{fig:vis_webui}
\end{figure*}

\subsection{In-Depth Analysis}
\label{sec:analysis}

We discuss several important questions about the design assumptions and practical implications of \methodname.

\paragraph{Q1: Is $z_0$ always the correct correction target for off-trajectory states?}
The correction loss directs the model toward the original training sample $z_0$ regardless of how far the rollout has drifted. In principle, an off-trajectory state $z_{\sigma_{t'}}$ could be legitimately denoised to a different clean sample $z_0'$ that is more consistent with its current latent structure. \methodname's assumption is that the deviation is small enough, bounded by a single ODE step, so that $z_0$ remains the most plausible target. This assumption is strongest at low noise levels, where the latent is already close to $z_0$ and the set of plausible clean endpoints is narrow. At high noise levels, the assumption is less obvious but still supported by two factors: (i)~the one-step rollout deviation is proportional to $1/K$, which is small relative to the total trajectory; (ii)~re-noising with the shared $z_1$ preserves proximity to the original transport ray (Appendix~\ref{app:soar_correction}). The deliberate choice of $K{=}1$ rollout step keeps the deviation bounded and is a key design safeguard.

\paragraph{Q2: Why does high-noise correction improve semantics but not aesthetics?}
GenEval and OCR (semantic accuracy) improve substantially, while Aesthetic Score and PickScore (perceptual quality) improve only modestly. This follows from a noise-regime analysis: semantic decisions such as object count, spatial arrangement, and text layout are made at high-noise stages where the latent is ambiguous, whereas low-noise stages refine texture and detail. \methodname's correction has the largest impact at high noise, where the off-trajectory deviation is largest. In effect, high-noise correction trains the model to reach the correct semantic structure from a wider range of initial noise samples $z_1$, functioning similarly to ``golden initial noise'' selection but at training time rather than inference time.

\paragraph{Q3: Does trajectory correction reduce generation diversity?}
Trajectory correction may reduce diversity at high-noise levels by narrowing the set of initial noise realizations that lead to semantically valid outputs. However, we argue that this is a desirable trade-off rather than a deficiency. Unconstrained diversity in diffusion models includes outputs with wrong object counts, malformed anatomy, and broken text, i.e., semantically invalid samples that no user wants. What matters in practice is \emph{diversity under semantic consistency}: the range of valid, high-quality outputs the model can produce for a given prompt. By teaching the denoiser to stay on-trajectory, \methodname eliminates the portion of the output space that corresponds to off-trajectory failures while preserving the legitimate variation encoded in the training distribution. In this sense, trajectory correction acts as a semantic consistency constraint that channels the model's capacity toward meaningful diversity rather than erratic noise-dependent failures. A thorough quantitative evaluation of this trade-off, with per-prompt LPIPS variance and distribution-level Vendi Score, is planned for future work.

\paragraph{Q4: How should the noise weighting $w(\sigma)$ be designed?}
The correction target $v_{\mathrm{corr}} = (z_{\sigma_{t'}} - z_0)/\sigma_{t'}$ has a denominator $\sigma_{t'}$ that becomes small at low noise levels, amplifying the gradient magnitude. The current implementation inherits the base flow matching weight $w(\sigma)$, which was designed for the standard velocity target and does not specifically account for this amplification. A dedicated weighting schedule, for example one that down-weights the correction loss at very low $\sigma$ to prevent gradient explosion or that up-weights mid-noise levels where the semantic-vs-detail trade-off is most active, could improve training stability and metric balance. We leave this investigation to future work.

\subsection{Qualitative Visualization}
\label{sec:qualitative}

Figures~\ref{fig:vis_aesth}--\ref{fig:vis_webui} present training-time visualizations across three data domains. The \emph{high-aesthetic} subset (Figure~\ref{fig:vis_aesth}) spans portraits, landscapes, and anime; as training progresses, \methodname produces increasingly coherent global compositions while preserving fine detail in faces, lighting, and textures. The \emph{poster} (Figure~\ref{fig:vis_poster}) and \emph{WebUI} (Figure~\ref{fig:vis_webui}) subsets test more structured generation tasks that demand precise layout, typographic placement, shadow rendering, and stylistic consistency. Across all three domains, \methodname demonstrates high learning efficiency and accuracy: the dense per-timestep correction signal enables the model to absorb not only semantic structure but also the subtle visual properties, including style, spatial hierarchy, specular effects, and drop shadows, that distinguish professional-quality outputs from generic generations. These are precisely the fine-grained attributes that a sparse terminal reward would struggle to credit to specific denoising steps, yet \methodname captures them through direct trajectory-level supervision.

%% file: sec/4_relatedwork.tex
\section{Related Work}

\subsection{Diffusion Models and Flow Matching}

Diffusion models generate data by learning to reverse a noise-corruption process~\citep{ddpm_ho_2020,song_scorebased_2021}. Latent diffusion~\citep{LDM_rombach_2022} moves this process into a compressed latent space, enabling high-resolution synthesis. Flow matching~\citep{lipman_flow_2023,liu_rectifiedflow_2023} reformulates generation as learning a velocity field along a linear interpolation between noise and data, yielding efficient ODE-based sampling. The rectified-flow variant~\citep{liu_rectifiedflow_2023} is adopted by current state-of-the-art systems including SD3.5~\citep{sd35_esser_2024} and FLUX~\citep{Flux_blacklabs_2025}, and the DiT architecture~\citep{dit_peebles_2023} has become the standard backbone. Our work builds on this flow-matching foundation and addresses a limitation of the training procedure, not the architecture.

\subsection{Post-Training for Diffusion Models}

Post-training methods can be broadly grouped into four categories.
\emph{Supervised fine-tuning} continues the flow matching or diffusion objective on curated data to improve specific capabilities such as text rendering or compositional accuracy. While simple and stable, SFT inherits the exposure-bias limitation discussed in Section~\ref{sec:paradigms}.
\emph{Reward-based optimization} uses differentiable reward models such as ImageReward~\citep{imagereward_xu_2023}, HPS~\citep{HPS_Wu_2023,HPSv2_Wu_2023}, and PickScore to directly backpropagate preference signals into the generator.
\emph{DPO-style methods}~\citep{wallace_diffusiondpo_2024} adapt direct preference optimization to diffusion models by constructing positive and negative sample pairs and adjusting their relative likelihoods. In the flow-matching setting, however, the paired samples typically lie on different interpolation paths, lacking the geometric structure that \methodname exploits. DiffusionNFT~\citep{zheng_diffusionnft_2026} extends this idea to the online setting by splitting model generations into positive and negative subsets according to reward and optimizing an implicit policy improvement direction on the forward process via flow matching, bypassing likelihood estimation entirely.
\emph{RL-style methods} treat denoising as an MDP and apply policy gradients. Flow-GRPO~\citep{flowgrpo_liu_2025} introduces GRPO to flow matching via an ODE-to-SDE conversion, achieving GenEval accuracy of 95\% on SD3.5-M. SimpleAR~\citep{wang2025simplear} and T2I-R1~\citep{jiang2025t2i} explore similar ideas for autoregressive and reasoning-augmented generators.
Our work is positioned earlier in the post-training stack: rather than using reward optimization as the first tool, \methodname directly repairs the generator's off-trajectory denoising behavior before any reward signal is introduced.

\subsection{Exposure Bias in Sequential Generation}

Exposure bias was first identified in autoregressive language modeling, where models are trained with teacher forcing but must condition on their own predictions at test time~\citep{bengio_scheduled_2015}. Scheduled sampling~\citep{bengio_scheduled_2015} and related curriculum strategies partially mitigate this gap by occasionally substituting model predictions for ground-truth tokens during training.

In diffusion models the same mismatch occurs along the denoising axis. \citet{ning_input_2023} show that adding small perturbations to the training input (``input perturbation'') improves sample quality by exposing the model to slightly off-distribution states. \citet{li_exposure_2024} provide a formal analysis of the exposure-bias effect in diffusion training and propose an auxiliary objective to reduce the compounding error. These works diagnose the problem and offer training-time regularization; \methodname goes further by constructing on-policy off-trajectory states from the model's own rollout and supervising the correction explicitly, providing a more direct remedy that naturally integrates into the post-training pipeline.

\subsection{External Correction and Reasoning-Based Control}

A complementary line of work improves generation quality through external feedback loops rather than weight updates. Self-correcting LLM-controlled diffusion~\citep{wu2024self}, Idea2Img~\citep{yang2024idea2img}, dynamic prompt optimization~\citep{hao2023optimizing,mo2024dynamic,manas2024improving}, reasoning-augmented re-prompting~\citep{reprompt_wu_2025,yang2024mastering}, and chain-of-thought prompt rewriting~\citep{wang2025promptenhancer} leverage multimodal language models to iteratively refine prompts or select among candidate outputs. These methods are valuable when the generator weights are frozen, but they do not modify the nearby off-trajectory regions visited during inference. \methodname is complementary: it targets the generator parameters directly, improving the base model that external correction methods build upon.

%% file: sec/5_conclusion.tex
\section{Conclusion}

We have proposed \methodname, a post-training method that replaces standard SFT as a stronger first stage directly following pretraining in the diffusion model stack. By identifying exposure bias, the distributional mismatch between ground-truth training states and model-induced inference states, as a unifying cause of many persistent generation failures, \methodname provides a direct remedy: it constructs off-trajectory states from the model's own rollout, re-noises them, and supervises the denoiser with an analytically derived correction target anchored to the original clean sample.

\methodname combines the desirable properties of both SFT and RL while avoiding their respective weaknesses. Like SFT, its supervision is dense and analytically derived, requiring no reward model. Like RL, it trains on states the model actually encounters, making it genuinely on-policy. Unlike RL, the correction signal acts at the exact timestep where the deviation occurs, eliminating the credit-assignment problem entirely. Experiments on SD3.5-Medium confirm that \methodname improves both rule-based and model-based metrics over SFT, achieves monotonic reward improvement on curated data subsets, and outperforms explicit RL in final metric value without access to a reward function.

Several directions remain open. First, although our results show uniform metric improvements, we have not yet evaluated the effect on generation \emph{diversity} (e.g., via LPIPS or Vendi Score); the trajectory-correction bias toward training data may narrow the output distribution, and understanding this trade-off is important. Second, the noise-dependent weight $w(\sigma)$ is currently inherited from the base flow matching schedule; a dedicated analysis of how to weight the correction loss across noise levels could further improve the method. Finally, the bias-correction principle is not specific to image generation: whenever a flow-based or diffusion generator follows a multi-step trajectory, the same mechanism applies. Extending \methodname to video generation, 3D synthesis, and model distillation is a natural next step.

%% file: sec/x_contributor.tex
\section{Acknowledgements}
We would like to thank  Zhenxi Li, Yixuan Shi, Tingting Gao for their valuable inputs and suggestions.
\bigskip



%% file: sec/a_appendix_diffusion_derivation.tex
\section{Derivation and Training Details}
\label{app:diffusion_derivation}

This appendix provides the formal derivation of \methodname from the conditional optimal-transport structure of rectified flow. We first establish the transport-ray geometry (A.1), formalize the exposure bias as a distributional shift (A.2), derive the correction target from the clean-endpoint condition (A.3), connect the per-sample loss to distribution matching (A.4), and give the full loss aggregation and detailed algorithm (A.5).

\subsection{Rectified Flow and Conditional Transport}
\label{app:cond_transport}

Rectified flow~\citep{liu_rectifiedflow_2023} learns a velocity field that generates a straight-line probability path from the noise distribution $p_1 = \mathcal{N}(0,I)$ to the data distribution $p_0 = q(z_0)$. Each training pair $(z_0, z_1)$ with $z_0 \sim q(z_0)$ and $z_1 \sim \mathcal{N}(0,I)$ defines a \emph{transport ray}, the linear path
\begin{equation}
z_t = (1-t)\,z_0 + t\,z_1, \qquad t \in [0,1].
\label{eq:app_ray}
\end{equation}
The \emph{conditional velocity} along this ray is constant:
\begin{equation}
u(z_t \mid z_0, z_1) = z_1 - z_0.
\label{eq:app_cond_vel}
\end{equation}
The flow matching objective trains $v_\theta(z_t, c, t)$ to approximate the \emph{marginal velocity}, i.e., the conditional expectation of $u$ given the noisy state:
\begin{equation}
u_t(z_t) = \mathbb{E}[z_1 - z_0 \mid z_t] = \frac{z_t - \mathbb{E}[z_0 \mid z_t]}{t}.
\label{eq:app_marginal_vel}
\end{equation}
The second equality follows from Eq.~\eqref{eq:app_ray}. The model's velocity prediction at any state $z$ and noise level $\sigma$ implies a \emph{clean-endpoint map}:
\begin{equation}
\hat{z}_0(z, \sigma) = z - \sigma \cdot v_\theta(z, c, \sigma).
\label{eq:app_clean_endpoint}
\end{equation}
This map is central to \methodname: both the on-trajectory and off-trajectory losses can be understood as requiring $\hat{z}_0$ to land on the correct clean sample $z_0$.

\subsection{Exposure Bias as Distributional Shift}
\label{app:exposure_bias}

During training, the model observes states drawn from the \emph{ground-truth} (on-trajectory) distribution:
\begin{equation}
p_\sigma^{\mathrm{data}}(z) = \int \delta\bigl((1{-}\sigma)z_0 + \sigma z_1\bigr)(z)\, q(z_0)\,\mathcal{N}(z_1)\,dz_0\,dz_1.
\label{eq:app_pdata}
\end{equation}
The SFT objective minimizes the velocity error under this distribution:
\begin{equation}
\mathcal{L}_{\mathrm{SFT}} = \mathbb{E}_{z \sim p_\sigma^{\mathrm{data}}}\bigl[\lVert v_\theta(z, \sigma) - u_\sigma(z)\rVert^2\bigr].
\end{equation}

During inference, the model solves the ODE $dz/d\sigma = v_\theta(z, \sigma)$ from $\sigma{=}1$ to $\sigma{=}0$. The resulting state at noise level $\sigma$ is
\begin{equation}
\hat{z}_\sigma = z_1 + \int_1^\sigma v_\theta(\hat{z}_s, c, s)\,ds,
\end{equation}
and its distribution $p_\sigma^\theta(z)$ is defined implicitly by this ODE. Since each state depends on the model's previous predictions, any velocity error at an earlier step shifts all subsequent states.

\paragraph{Error accumulation.}
Consider a single Euler step from $\sigma_0$ to $\sigma_1 = \sigma_0 + \Delta\sigma$ (with $\Delta\sigma < 0$ during denoising). Let $\epsilon(\sigma_0) = v_\theta(z_{\sigma_0}, \sigma_0) - u_{\sigma_0}(z_{\sigma_0})$ denote the velocity error. The state deviation after one step is
\begin{equation}
\Delta z = |\Delta\sigma| \cdot \lVert\epsilon(\sigma_0)\rVert.
\end{equation}
Over $K$ Euler steps, the total deviation satisfies
\begin{equation}
\lVert \hat{z}_0 - z_0^{\mathrm{ideal}} \rVert \le \sum_{k=1}^{K} \frac{1}{K}\lVert\epsilon(\sigma_k)\rVert + \text{(higher-order cross terms)}.
\label{eq:app_error_bound}
\end{equation}
Even if each $\lVert\epsilon\rVert$ is small, the sum can become large because errors at early (high-noise) steps also shift the states at which later predictions are evaluated, creating a compounding effect. This is precisely the exposure-bias phenomenon: the model is trained on $p_\sigma^{\mathrm{data}}$ but evaluated on $p_\sigma^\theta$, and the gap grows with the number of sampling steps.

\subsection{SOAR Correction via Conditional Optimal Transport}
\label{app:soar_correction}

\methodname addresses the exposure bias by expanding the training distribution from $p^{\mathrm{data}}$ to $p^{\mathrm{data}} \cup p^{\mathrm{rollout}}$, where $p^{\mathrm{rollout}}$ consists of states the model actually generates during a short rollout. We now formalize the construction and the correction target.

\subsubsection{Off-Trajectory State Construction}

Starting from a training pair $(z_0, z_1, c)$ and the on-trajectory state $z_{\sigma_{t_0}} = (1{-}\sigma_{t_0})z_0 + \sigma_{t_0}z_1$, we compute a stop-gradient CFG rollout velocity:
\begin{equation}
v_{\mathrm{cfg}} = \operatorname{sg}\bigl[v_\theta(z_{\sigma_{t_0}}, \varnothing, t_0) + w_{\mathrm{cfg}}\bigl(v_\theta(z_{\sigma_{t_0}}, c, t_0) - v_\theta(z_{\sigma_{t_0}}, \varnothing, t_0)\bigr)\bigr],
\end{equation}
where $\varnothing$ denotes the null condition and $\operatorname{sg}[\cdot]$ prevents gradient flow. A single ODE step produces
\begin{equation}
\hat{z}_{\sigma_{t_1}} = z_{\sigma_{t_0}} + (\sigma_{t_1} - \sigma_{t_0})\,v_{\mathrm{cfg}}, \qquad t_1 = \max(t_0 - 1/K,\, 0).
\label{eq:app_ode_step}
\end{equation}
The state $\hat{z}_{\sigma_{t_1}}$ deviates from the ideal $z_{\sigma_{t_1}}^{\mathrm{ideal}} = (1{-}\sigma_{t_1})z_0 + \sigma_{t_1}z_1$ by
\begin{equation}
\delta = \hat{z}_{\sigma_{t_1}} - z_{\sigma_{t_1}}^{\mathrm{ideal}} = (\sigma_{t_1} - \sigma_{t_0})\bigl(v_{\mathrm{cfg}} - (z_1 - z_0)\bigr).
\label{eq:app_deviation}
\end{equation}
This deviation is proportional to $|\sigma_{t_1} - \sigma_{t_0}| = 1/K$ and to the CFG-adjusted velocity error, both of which are bounded in practice.

We then construct auxiliary states by re-noising $\hat{z}_{\sigma_{t_1}}$ toward the noise endpoint $z_1$. For an auxiliary noise level $\sigma_{t'} \in [\sigma_{t_1}, 1]$:
\begin{align}
\alpha &= \frac{\sigma_{t'} - \sigma_{t_1}}{1 - \sigma_{t_1}}, \label{eq:app_alpha}\\
z_{\sigma_{t'}} &= (1-\alpha)\,\hat{z}_{\sigma_{t_1}} + \alpha\, z_1. \label{eq:app_aux_state}
\end{align}

\subsubsection{Geometric Insight: Same-Noise Re-Noising}

A critical design choice is that re-noising uses the \textbf{same} $z_1$ as the forward process, rather than drawing fresh Gaussian noise. This has a precise geometric consequence.

The ideal state at noise level $\sigma_{t'}$ on the original transport ray is $z_{\sigma_{t'}}^{\mathrm{ideal}} = (1{-}\sigma_{t'})z_0 + \sigma_{t'}z_1$. Writing the auxiliary state from Eq.~\eqref{eq:app_aux_state} and substituting the deviation $\delta$ from Eq.~\eqref{eq:app_deviation}:
\begin{align}
z_{\sigma_{t'}} &= (1-\alpha)\bigl(z_{\sigma_{t_1}}^{\mathrm{ideal}} + \delta\bigr) + \alpha\, z_1 \notag \\
&= \underbrace{(1-\alpha)\,z_{\sigma_{t_1}}^{\mathrm{ideal}} + \alpha\, z_1}_{\text{ideal state at }\sigma_{t'}} + (1-\alpha)\,\delta.
\label{eq:app_aux_deviation}
\end{align}
Therefore:
\begin{equation}
\lVert z_{\sigma_{t'}} - z_{\sigma_{t'}}^{\mathrm{ideal}} \rVert = (1-\alpha)\,\lVert\delta\rVert.
\label{eq:app_bounded_dev}
\end{equation}

This bound has two important implications:
\begin{enumerate}[leftmargin=2em,itemsep=2pt]
\item The deviation from the ideal ray is \emph{strictly bounded} by the one-step rollout error $\lVert\delta\rVert$ and \emph{shrinks} as $\alpha \to 1$ (i.e., as the auxiliary noise level approaches pure noise). At $\alpha = 1$, the auxiliary state coincides with $z_1$ regardless of the rollout error.
\item Because the auxiliary state remains close to the transport ray $(z_0 \leftrightarrow z_1)$, the original $z_0$ is still the uniquely correct clean-endpoint target under the learned coupling. The state has not drifted to a different mode of the data distribution where a different $z_0'$ would be more appropriate.
\end{enumerate}

If instead we used fresh noise $z_1' \sim \mathcal{N}(0,I)$, the auxiliary state would be $z_{\sigma_{t'}}' = (1{-}\alpha)\hat{z}_{\sigma_{t_1}} + \alpha z_1'$. This state lies on a \emph{different} interpolation path $(z_0 \leftrightarrow z_1')$, and the ideal target from the perspective of this path would involve $z_1'$, not the original $z_1$. Using $z_0$ as the correction anchor for such a state introduces a systematic bias that grows with $\alpha$ and with the distance $\lVert z_1 - z_1'\rVert$. The ablation in Section~\ref{sec:ablation} confirms that shared $z_1$ outperforms fresh noise on rule-based metrics.

\subsubsection{Deriving the Correction Target}

We derive the correction target following the same logic as Section~\ref{sec:correction}, with full algebraic detail.

\paragraph{From local trajectory matching to goal consistency.}
The most direct way to correct a drifted state is to require that, after a small step $\Delta t$, the off-trajectory state $z_{\sigma}'$ reconverges with the on-trajectory state $z_\sigma$:
\begin{equation}
z_\sigma + \Delta t \cdot v_{\mathrm{gt}} = z_{\sigma}' + \Delta t \cdot v_{\mathrm{corr}}.
\label{eq:app_local_match}
\end{equation}
However, the step size $\Delta t$ introduces an ambiguity: different choices yield different correction targets, and the optimal $\Delta t$ depends on the downstream trajectory. We resolve this by requiring both states to reach the same final endpoint $z_0$, eliminating $\Delta t$ entirely. On the ideal trajectory, $z_\sigma = (1{-}\sigma)z_0 + \sigma z_1$ with $v_{\mathrm{gt}} = z_1 - z_0$ satisfies
\begin{equation}
z_\sigma - \sigma \cdot v_{\mathrm{gt}} = z_0.
\label{eq:app_on_goal}
\end{equation}
Imposing the same goal consistency condition on the off-trajectory state gives
\begin{equation}
z_{\sigma}' - \sigma \cdot v_{\mathrm{corr}} = z_0,
\label{eq:app_off_goal}
\end{equation}
which yields the closed-form correction target:
\begin{equation}
v_{\mathrm{corr}} = \frac{z_{\sigma}' - z_0}{\sigma}.
\label{eq:app_vstar}
\end{equation}

\paragraph{On-trajectory verification.}
When $z_{\sigma}' = z_\sigma$ (no deviation), substituting $z_\sigma = (1{-}\sigma)z_0 + \sigma z_1$ into Eq.~\eqref{eq:app_vstar}:
\begin{equation}
v_{\mathrm{corr}} = \frac{(1{-}\sigma)z_0 + \sigma z_1 - z_0}{\sigma} = z_1 - z_0 = v_{\mathrm{gt}}.
\end{equation}
This recovers the standard flow matching target, confirming that the correction objective generalizes the SFT loss.

\paragraph{Off-trajectory case.}
For an auxiliary state $z_{\sigma_{t'}}$ from Eq.~\eqref{eq:app_aux_state}, which deviates from the ideal ray by $(1{-}\alpha)\delta$ (Eq.~\ref{eq:app_aux_deviation}), the correction target differs from $v_{\mathrm{gt}}$ by:
\begin{equation}
v_{\mathrm{corr}} - v_{\mathrm{gt}} = \frac{z_{\sigma_{t'}} - z_{\sigma_{t'}}^{\mathrm{ideal}}}{\sigma_{t'}} = \frac{(1-\alpha)\,\delta}{\sigma_{t'}}.
\label{eq:app_vcorr_deviation}
\end{equation}
This correction term is proportional to the rollout deviation $\delta$, scaled by the interpolation factor $(1{-}\alpha)$ and the noise level $\sigma_{t'}$. It is the dense, per-timestep signal that steers the model back toward $z_0$ from the off-trajectory states it actually visits during inference.

\subsection{Distribution Matching Interpretation}
\label{app:dist_matching}

We can frame the combined \methodname objective in distributional terms. For a given training pair $(z_0, z_1)$, the model's clean-endpoint prediction at state $z$ with noise level $\sigma$ is a point mass:
\begin{equation}
\mu_\theta(z, \sigma) = \delta_{\hat{z}_0(z, \sigma)} = \delta_{z - \sigma v_\theta(z,\sigma)}.
\end{equation}
The target is also a point mass: $\mu^* = \delta_{z_0}$. The 2-Wasserstein distance between them is:
\begin{equation}
W_2^2(\mu_\theta, \mu^*) = \lVert \hat{z}_0(z,\sigma) - z_0 \rVert^2 = \sigma^2 \lVert v_\theta(z,\sigma) - v^*(z,\sigma) \rVert^2.
\label{eq:app_w2}
\end{equation}

The SFT objective minimizes $W_2^2$ in expectation over $z \sim p_\sigma^{\mathrm{data}}$:
\begin{equation}
\mathcal{L}_{\mathrm{SFT}} = \mathbb{E}_{z_0,z_1,\sigma}\bigl[W_2^2\bigl(\mu_\theta(z_\sigma^{\mathrm{data}}, \sigma),\, \delta_{z_0}\bigr)\bigr].
\end{equation}

The \methodname objective extends this to both on-trajectory and off-trajectory states:
\begin{equation}
\mathcal{L}_{\mathrm{SOAR}} = \mathbb{E}_{z_0,z_1,\sigma}\bigl[
  \underbrace{W_2^2(\mu_\theta(z_\sigma^{\mathrm{data}}, \sigma), \delta_{z_0})}_{\text{on-trajectory}}
  + \lambda \underbrace{W_2^2(\mu_\theta(z_\sigma^{\mathrm{roll}}, \sigma), \delta_{z_0})}_{\text{off-trajectory}}
\bigr],
\end{equation}
where $z_\sigma^{\mathrm{roll}}$ denotes an auxiliary state from Eq.~\eqref{eq:app_aux_state}. In practice, the $\sigma^2$ factor from Eq.~\eqref{eq:app_w2} is absorbed into the noise-dependent weight $w(\sigma)$, and the expectation is approximated by batch-level summation.

The key difference from SFT is that the expectation now covers $p^{\mathrm{data}} \cup p^{\mathrm{rollout}}$ instead of $p^{\mathrm{data}}$ alone. By training on states the model actually visits during inference, \methodname directly closes the exposure-bias gap identified in Section~\ref{app:exposure_bias}.

\subsection{Loss Aggregation and Practical Details}
\label{app:aggregation}

For a batch of size $B$ with $P$ valid auxiliary points in total, let $w(\sigma)$ denote the noise-dependent loss weight inherited from the base flow matching schedule. The per-sample on-trajectory and off-trajectory losses are:
\begin{equation}
\ell_{\mathrm{base}}^{(b)} = w(\sigma_{t_0}^{(b)})\lVert v_\theta(z_{\sigma_{t_0}}^{(b)}, c^{(b)}, t_0^{(b)}) - v_{\mathrm{gt}}^{(b)}\rVert^2, \qquad
\ell_{\mathrm{aux}}^{(p)} = w(\sigma_{t'}^{(p)})\lVert v_\theta(z_{\sigma_{t'}}^{(p)}, c^{(p)}, t'^{(p)}) - v_{\mathrm{corr}}^{(p)}\rVert^2,
\end{equation}
where both norms are averaged over the latent dimension $d$. The combined objective aggregates all supervised items and normalizes once:
\begin{equation}
\mathcal{L}_{\mathrm{total}} = \frac{\displaystyle\sum_{b=1}^{B}\ell_{\mathrm{base}}^{(b)} + \lambda\sum_{p=1}^{P}\ell_{\mathrm{aux}}^{(p)}}{B + \lambda P}.
\end{equation}
Under distributed training, $B$ and $P$ are synchronized across workers via all-reduce before normalization to ensure consistent gradient scaling.

\begin{algorithm}[t]
\caption{Detailed SOAR training procedure (with optional SDE branches)}
\label{alg:soar_detailed}
\begin{algorithmic}[1]
\Require velocity field $v_\theta$, condition $c$, correction weight $\lambda$
\Statex \hspace{\algorithmicindent} rollout CFG scale $w_{\mathrm{cfg}}$, ODE/SDE branch count $M$
\Statex \hspace{\algorithmicindent} auxiliary points per branch $N$, rollout step count $K$
\For{each training batch}
    \State Sample clean latent $z_0$, Gaussian noise $z_1 \sim \mathcal{N}(0,I)$, and training time $t_0$
    \State Compute $\sigma_{t_0}$ and construct $z_{\sigma_{t_0}} = (1-\sigma_{t_0}) z_0 + \sigma_{t_0} z_1$
    \State $v_{\mathrm{on}} \gets v_\theta(z_{\sigma_{t_0}}, c, t_0)$;\quad $v_{\mathrm{gt}} \gets z_1 - z_0$
    \State $\mathcal{L}_{\mathrm{main}} \mathrel{+}= w(\sigma_{t_0})\lVert v_{\mathrm{on}} - v_{\mathrm{gt}}\rVert^2$;\quad $N_{\mathrm{main}} \gets B$;\quad $N_{\mathrm{aux}} \gets 0$
    \If{$\lambda > 0$ and $N > 0$}
        \State Compute $v_{\mathrm{cfg}} \gets \operatorname{sg}[v_{\mathrm{uncond}} + w_{\mathrm{cfg}}(v_{\mathrm{cond}} - v_{\mathrm{uncond}})]$
        \State $t_1 \gets \max(t_0 - 1/K, 0)$;\quad $\sigma_{t_1} \gets \sigma(t_1)$
        \State \textbf{Path 0 (ODE):} $\hat{z}_{\sigma_{t_1}}^{(0)} \gets z_{\sigma_{t_0}} + (\sigma_{t_1} - \sigma_{t_0})\,v_{\mathrm{cfg}}$
        \If{$M > 1$ and $t_1 > 0$} \Comment{optional SDE branches}
            \For{$m = 1, \dots, M{-}1$}
                \State $\hat{z}_{\sigma_{t_1}}^{(m)} \gets \Psi(z_{\sigma_{t_0}}, v_{\mathrm{cfg}}, \sigma_{t_0}, \sigma_{t_1}; \eta)$
            \EndFor
        \EndIf
        \For{each branch endpoint $\hat{z}_{\sigma_{t_1}}^{(m)}$, $m = 0, \dots, M'{-}1$}
            \For{$n = 1, \dots, N$}
                \State Sample $\sigma_{t'} \sim \mathrm{Uniform}[\sigma_{t_1}, 1]$
                \State $\alpha \gets (\sigma_{t'} - \sigma_{t_1})/(1 - \sigma_{t_1})$
                \State $z_{\sigma_{t'}} \gets (1{-}\alpha)\,\hat{z}_{\sigma_{t_1}}^{(m)} + \alpha\, z_1$
                \State $v_{\mathrm{off}} \gets v_\theta(z_{\sigma_{t'}}, c, t')$;\quad $v_{\mathrm{corr}} \gets (z_{\sigma_{t'}} - z_0)/\sigma_{t'}$
                \State $\mathcal{L}_{\mathrm{aux}} \mathrel{+}= \lambda\,w(\sigma_{t'})\lVert v_{\mathrm{off}} - v_{\mathrm{corr}}\rVert^2$;\quad $N_{\mathrm{aux}} \mathrel{+}= |\text{batch}|$
            \EndFor
        \EndFor
    \EndIf
    \State All-reduce $N_{\mathrm{main}}, N_{\mathrm{aux}}$ across workers
    \State $\mathcal{L}_{\mathrm{total}} \gets (\mathcal{L}_{\mathrm{main}} + \mathcal{L}_{\mathrm{aux}}) / (N_{\mathrm{main}} + \lambda\, N_{\mathrm{aux}})$
    \State Update $\theta$ using $\nabla_\theta \mathcal{L}_{\mathrm{total}}$
\EndFor
\end{algorithmic}
\end{algorithm}

In the default configuration ($M{=}1$, ODE-only), the algorithm reduces to Algorithm~\ref{alg:soar} in the main text. The optional SDE branches ($M > 1$) apply a stochastic one-step operator $\Psi$ (e.g., coefficient-preserving sampling with noise scale $\eta$) and are evaluated in the ablation study (Section~\ref{sec:ablation}). In either case, the core supervision pattern is the same: the model is trained on both the ideal ground-truth state and the nearby off-trajectory states that its own rollout can produce.